\pdfoutput=1

\documentclass[11pt]{article}

\usepackage{acl}

\usepackage{times}
\usepackage{latexsym}
\usepackage{makecell}
\usepackage{amsfonts}

\newcommand{\qual}{\raisebox{-2.4pt}{\includegraphics[scale=0.07]{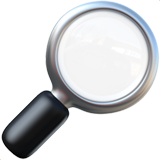}}}

\usepackage[T1]{fontenc}

\usepackage[utf8]{inputenc}

\usepackage{microtype}

\usepackage{graphicx} 
\usepackage{multirow} 
\usepackage{booktabs} 
\usepackage{multirow} 
\usepackage{subcaption} 
\newcommand{\best}[1]{\colorbox{black!10}{#1}}

\usepackage{tikz}
\newlength\barheight 
\newlength\barwidth 

\newcommand\blfootnote[1]{%
  \begingroup
  \renewcommand\thefootnote{}\footnote{#1}%
  \addtocounter{footnote}{-1}%
  \endgroup
}

\newcommand{\positive}{\textbf{\textcolor{forestgreen}{goal}}}
\newcommand{\negative}{\textbf{\textcolor{red}{avoid}}}

\usepackage{makecell}
\usepackage{listings}
\usepackage{times,latexsym}
\usepackage{url}
\usepackage{color,soul} 
\usepackage[T1]{fontenc}
\usepackage{amsmath}

\usepackage{array,booktabs,ragged2e}
\usepackage{multirow}
\newcolumntype{R}[1]{>{\RaggedLeft\arraybackslash}p{#1}}
\newcolumntype{L}[1]{>{\RaggedRight\arraybackslash}m{#1}}
\usepackage{pifont} 

\newcommand{\cmark}{\ding{51}}
\newcommand{\xmark}{\ding{55}}

\newcommand{\cluegiver}{\textsc{The Clue Giver}}
\newcommand{\guesser}{\textsc{The Guesser}}
\newcommand{\dataset}{\textsc{Cultural Codes}}

\usepackage{subcaption}

\usepackage{amsmath, balance}
\usepackage{caption}
\usepackage{subcaption}
\usepackage{booktabs}
\usepackage{soul}

\definecolor{forestgreen}{HTML}{397727}

\title{Modeling Cross-Cultural Pragmatic Inference with Codenames Duet}

\newcommand\coauth{$^\star$}
\newcommand{\gt}{$^\ddag$}
\newcommand{\stanf}{$^\dagger$}
\newcommand{\usc}{$^\diamond$}
\author{Omar Shaikh\coauth \stanf  \hspace{0.8em}
        Caleb Ziems\coauth \stanf  \hspace{0.8em}
        William Held \gt \hspace{0.8em}
        Aryan J. Pariani \gt \hspace{0.8em}\\
        \textbf{Fred Morstatter} \usc \hspace{0.8em}
        \textbf{Diyi Yang} \stanf\\
         \stanf Stanford University, \gt Georgia Institute of Technology, \usc USC Information Sciences Institute\\
         \texttt{\small \{oshaikh, cziems, diyiy\}@stanford.edu}
        \texttt{\small\{wheld3, apariani3\}@gatech.edu}
        \texttt{\small fred@isi.edu}}

\date{}

\begin{document}
\maketitle

\begin{abstract}
Pragmatic reference enables efficient interpersonal communication.  Prior work uses simple reference games to test models of pragmatic reasoning, often with unidentified speakers and listeners. In practice, however, speakers’ sociocultural background shapes their pragmatic assumptions. For example, readers of this paper assume NLP refers to ``Natural Language Processing,'' and \textit{not} ``Neuro-linguistic Programming.'' This work introduces the \dataset{} dataset, which operationalizes sociocultural pragmatic inference in a simple word reference game. 

\dataset{} is based on the multi-turn collaborative two-player game, \textit{Codenames Duet}. Our dataset consists of 794 games with 7,703 turns, distributed across 153 unique players. Alongside gameplay, we collect information about players’ personalities, values, and demographics. Utilizing theories of communication and pragmatics, we predict each player’s actions via joint modeling of their sociocultural priors and the game context. Our experiments show that accounting for background characteristics significantly improves model performance for tasks related to both clue giving and guessing, indicating that sociocultural priors play a vital role in gameplay decisions.
\end{abstract}

\blfootnote{\coauth Equal contribution.}

\section{Introduction}
\begin{quote}
    {
    ``\textit{Most of our misunderstandings of other people are not due to any inability to... understand their words... [but that] we so often fail to understand a speaker's \textbf{intention}.}'' \\\phantom{abc}--- \textbf{George Armitage Miller} (\citeyear{miller1974psychology})
    }
\end{quote}
Certain pragmatic inferences can only be interpreted by individuals with shared backgrounds. For example, what researchers call \textit{fun} may not be \textit{fun} for kindergartners.
Theories from sociolinguistics, pragmatics, and communication aim to explain how sociocultual background affects interpersonal interaction~\citep{schramm1954communication}---especially since variation occurs across several dimensions: class~\citep{bernstein2003class, thomas1983cross}, age~\citep{labov2011principles}, gender~\citep{eckert2013language}, race~\citep{green2002african}, and more.

\begin{figure}[t!]
    \centering
    \includegraphics[width=\linewidth]{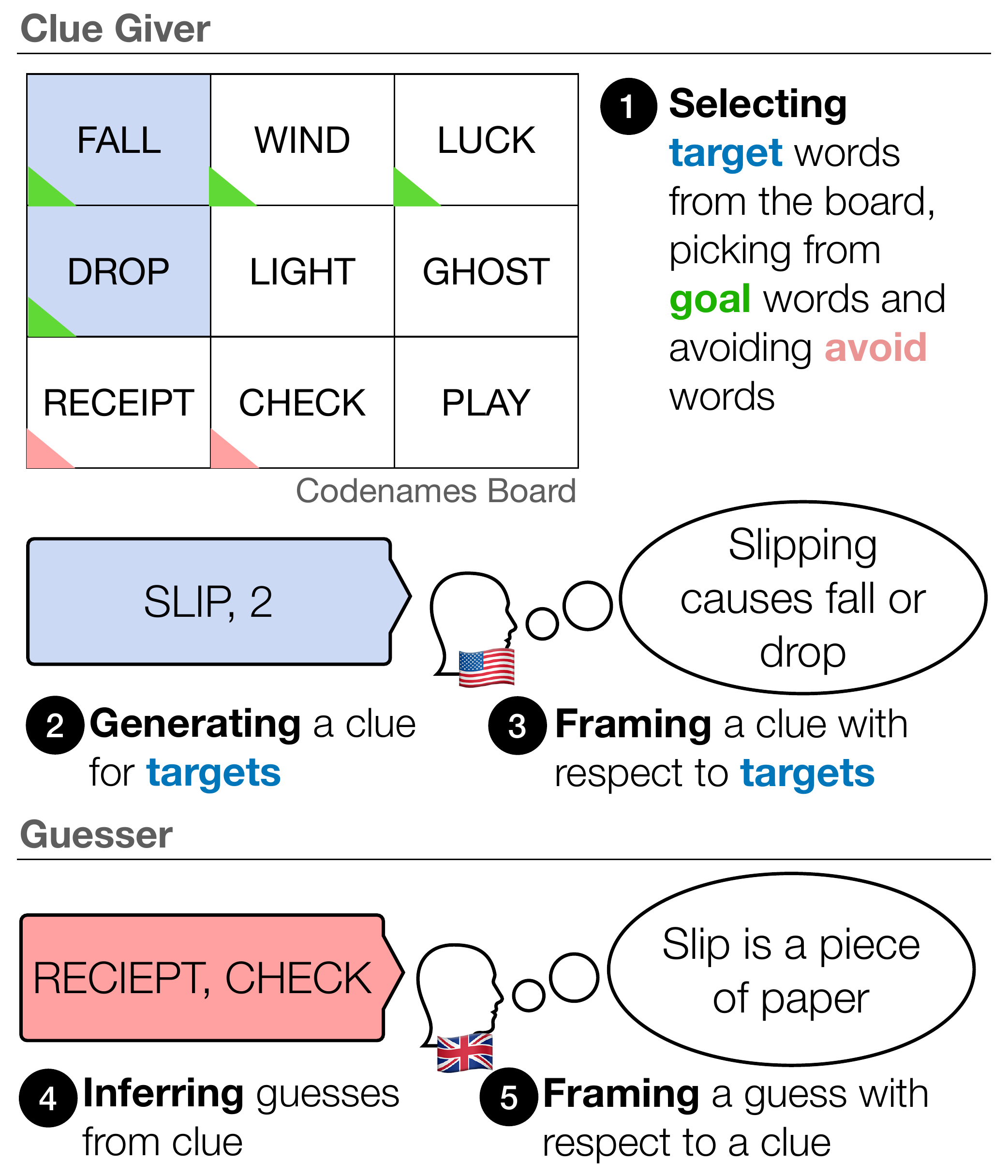}
    \caption{\textbf{An example interaction where difference in sociocultural background results in misinterpretation.} Steps 1-5 outline high-level gameplay tasks. \cluegiver{} targets the words \textit{fall} and  \textit{drop}, giving the hint \textit{slip}. \guesser{} misinterprets \textit{slip} as a piece of paper, guessing \textit{reciept} and \textit{check}. }
    \label{fig:first_fig}
\end{figure}

Rigorously modeling how culture affects pragmatic inference on \textit{all} axes is understandably challenging. The board game \textit{Codenames Duet} offers a more restricted setting of turn-based word reference between two players. In each round, \cluegiver{} provides a single-word clue; then \guesser{} must interpret this clue to select the intended word references on the game board. Ideal inferences come from the players' common ground---the set of shared beliefs between them~\citep{clark1996using}.
In practice, however, a player's behavior can be idiosyncratic. Each player has knowledge and experience that shape how they interpret clues and make guesses. 
When players' backgrounds differ, they may be more likely to misinterpret their partner, as seen in Figure \ref{fig:first_fig}. 

Inspired by the above, we model the role of sociocultural factors in pragmatic inference with a new task and a series of ablation experiments. First, we describe the \dataset{} dataset of cross-cultural \textit{Codenames Duet} gameplay, with relevant background information from the players' demographics, personalities, and political and moral values (\S \ref{the_dataset}). 
Then, we deconstruct each action in a game into a distinct modeling task, taking inspiration from work on cross-cultural pragmatics (\S \ref{all_tasks}). Finally, we model each task with/without sociocultural priors, and highlight how player background improves model performance (\S \ref{results}). Our dataset and code is released publicly at \url{https://github.com/SALT-NLP/codenames}
\section{Related Work}

\paragraph{Cross-Cultural Pragmatics and NLP} Pragmatics describes the nonliteral meaning that comes from context and social inference \citep{purpura2004assessing,thomas1983cross,hatch1992discourse}. Although some pragmatic categories are largely universal (e.g., politeness), they can be expressed differently in different sociocultural contexts \citep{taguchi2012context,shoshana1989cross,gudykunst1984communicating}. When an intended meaning is misinterpreted, this is known as `pragmatic failure' \citep{thomas1983cross}, and it is often the result of misaligned reference frames or differences in the common ground \citep{stadler2012cross,crawford2017new}. One axis of difference is between low and high-context cultures \citep{hofstede2001culture}, where high-context cultures rely more on shared background. Pragmatics also differs by age \citep{saryazdi2022pragmatic}, region, ethnicity, politics, and class \citep{thomas1983cross}, as does theory of mind reasoning \citep{fiske1979person,miller1984culture,shweder1984anthropology,lillard1998ethnopsychologies, lillard1999developing}. 

Outside of work on politeness \citep{sperlich2016interaction,fu2020facilitating}, sarcasm \citep{joshi2016cultural}, and irony \citep{karoui-etal-2017-exploring}, the NLP subfield has not closely considered cross-cultural pragmatics. While there is work on understanding the role of individual culture---for example, learning demographic word vectors~\citep{garimella2017demographic}, identifying deception/depression~\citep{soldner-etal-2019-box, loveys-etal-2018-cross}, or improving translation~\citep{specia-etal-2016-shared}---modeling \textbf{cross}-cultural pragmatic inference in communication remains a challenge \citep{hershcovich2022challenges}. 

Still, a culture-free pragmatics has played a central role in various NLP tasks, from instruction-following \citep{fried-etal-2018-unified}, image captioning \citep{andreas2016reasoning}, persona-consistent dialogue \citep{kim2020will}, and summarization \citep{shen-etal-2019-pragmatically}. Much of this work is grounded in Bayesian models of cognition \citep{griffiths2008bayesian}, with models like \textit{Bayesian Teaching} \citep{eaves2016infant}, \textit{Naive Utility Calculus} \citep{jara2016naive,jern2017people}, and the \textit{Rational Speech Acts} (RSA) model \citep{goodman2016pragmatic,franke2016probabilistic} that integrate language, world knowledge, and context to explain ideal pragmatic reasoning \citep{noveck2018experimental} and grounded reference \citep{monroe2017colors}. Instead of modeling socioculture in isolation, we model pragmatic inference, highlighting the role of culture in general interpersonal interaction.

\paragraph{Games as Testbeds for AI}
A significant body of work focuses on modeling optimal \textit{strategy} across a wide set of games, including Go~\citep{silver2016mastering}, Chess~\citep{schrittwieser2020mastering}, Poker~\citep{brown2017libratus}, Diplomacy~\citep{meta2022human}, D\&D~\citep{callison2022dungeons, zhou2022ai}, and Mafia \citep{ibraheem-etal-2022-putting}. Reference games are growing in popularity as testbeds for AI. Tests for artificial pragmatic reasoning often rely on sequential language games, where two players leverage private knowledge either to compete \citet{yao2021adversarial} or coordinate towards a common goal \citep{potts2012goal,khani2018planning,hawkins2015you}. In this vein, recent works have considered \textit{Codenames} \citep{koyyalagunta2021playing,kim2019cooperation,jaramillo2020word}, \textit{Connector} \citep{ashok2021contextual,kumar2021semantic,kovacs2022fast} \textit{InfoJigsaw} \citep{khani2018planning}, and image-based games \citep{bao2022learning}. Word association games have been used in psychology to study semantic associations in cultural \citep{korshuk2007learning} and religious \citep{tikhonova2014linguistic} contexts. We utilize games to model the effect of \textit{cross-cultural} interactions on pragmatic inference. 

\section{The \dataset{} Dataset}
\label{the_dataset}
This study has been approved by the Institutional Review Board (IRB) at the authors' institution.
The purpose of the \dataset{} dataset is to understand how measurable social factors influence dyadic communication \textit{in English}. By collecting relevant participant background information, we aim to understand how these factors affect linguistic reasoning in a collaborative reference game. 

\subsection{\textit{Codenames Duet} Game Overview}
\label{game_rules}
\textit{Codenames Duet} is a {collaborative} variant of \textit{Codenames} \citep{codenames} designed for 2 players. The players share a $5\times5$ board of $25$ common words. Each player has a distinct (but sometimes partially overlapping) map from words on the board to the following objectives: \positive{}, \textbf{neutral}, and \negative{}. One player's map is hidden from the opposing player. The objective of the game is for both players to guess all of their partner's \positive{} words  without guessing any of their partner's \negative{} words, as doing so results in an immediate loss. 

\dataset{} uses an adapted version of \textit{Codenames Duet}. With each turn, players alternate between the \cluegiver{} and \guesser{} roles. To begin the turn, \cluegiver{} \textbf{(1)} selects one or more associated \positive{} words as targets. Next, \cluegiver{} \textbf{(2)} provides a single word clue that relates to the associated target(s). This clue is displayed to \guesser{}, along with the number of targets she should find. The \cluegiver{} also \textbf{(3)} provides a justifying \textit{rationale} for the clue, describing the relationship between the clue and the target(s). This \textit{rationale} is not displayed to the partner. 
Using the clue and the number of target words \guesser{} \textbf{(4)} guesses targeted words. For each guess, \guesser{} \textbf{(5)} provides a justifying \textit{rationale} for the guess. After ending the turn, players alternate roles and continue until all \positive{} words are selected for both sides, or players are eliminated for guessing an \negative{} word. An overview of roles is illustrated in Figure \ref{fig:first_fig}. In \S\ref{all_tasks}, we formalize actions \textbf{(1)-(4)} as distinct modeling tasks.

\subsection{Selecting Board Game Words}
All experiments are run on a strategically filtered subset of the {400} words from \textit{Codenames Duet}. We select the {100} most abstract and semantically ambiguous board game words to elicit diverse responses from players. Since the \textit{polysemy} \citep{ravin2000polysemy} of a word---the number of related senses it includes---predicts the expected diversity of player responses, we retain only nouns with two or more senses in WordNet \citep{miller1995wordnet}. Next, we rank polysemous words with \citet{brysbaert2014concreteness}'s concreteness list, selecting the \textbf{100 most abstract words} according to the mean of their human concreteness scores (finalized list can be found in Appendix \ref{word_list}.) 

When a player starts a game, we initialize the board with a random subset of 25 words from the filtered 100. For each player, 9 words are randomly mapped to \positive{}, 3 are \negative{}, and 13 are \textbf{neutral}.

\subsection{Gameplay Data}
\label{gameplay_data}
To collect gameplay data, we modified an open-source implementation of \textit{Codenames Duet},\footnote{\url{https://github.com/jbowens/codenamesgreen}} automatically pairing individuals who visited the game website. To source players, we relied on Amazon's Mechanical Turk. We provided MTurkers with an initial instruction video detailing rules and how to play. To be eligible for the task, Turkers had to get $\geq80$\% questions right on a qualifying quiz about Codenames rules and gameplay (Appendix~\ref{appdx:qualification_test}). Average game length was around 17.4 minutes, and MTurkers were paid \$2.50 for every game. 

\paragraph{Gameplay Attributes} For each completed turn, we collected the following game state information from \cluegiver{}. Elements marked in \textcolor{gray}{gray} were hidden from \guesser{}. 
\begin{enumerate}
    \setlength\itemsep{0em}
    \item[] \textbf{Clue:} \cluegiver{'s }clue $c$ (e.g. $c$ could be ``\textsl{transport}'' for the target ``\textsl{car}'').  
    \item[] \textcolor{gray}{\textbf{Target Word(s):} (Hidden)} The target words $t_n$ (e.g. ``\textsl{car}'')
    that \cluegiver{} intended \guesser{} to guess.
    \item[] \textcolor{gray}{\textbf{Target Word(s) Rationale(s):} (Hidden)} A free-text phrase $r_n$, that describes the relationship between each target word $t_n$ and the clue $c$ (e.g. ``\textsl{a car is a mode of transport}'').
\end{enumerate}
To summarize, each turn from \cluegiver{} results in a clue $c$ and at least one target-rationale pair $(t_n, r_n)$. On the other hand, we collect the following for \textbf{\guesser{}}.
\begin{enumerate}
    \setlength\itemsep{0em}
    \item[] \textbf{Guesses}: The guesses $g_n$ that \guesser{} selected for \cluegiver{}'s clue $c$.
    \item[] \textcolor{gray}{\textbf{Rationale for Each Guess}}: A free-text phrase $r_n$ that relates the guess $g_n$ to the clue $c$
\end{enumerate}

Manual inspection revealed a wide range of rationales. To prevent models from exploiting variance, we instructed GPT-3 to normalize text, removing pronouns and determiners.\footnote{We use the text-davinci-003 variant from OpenAI. Without GPT-3 normalization, we find that model performance is artificially inflated.} We provided few-shot examples of reformatted rationales and manually inspected normalized outputs. Additional preprocessing information can be found in Appendix \ref{gpt3reformat}.

\subsection{Sociocultural Priors and Worker Diversity}
\label{demographic_info}
Because we aim to understand the role of sociocultural priors on gameplay, we asked Turkers to complete the standardized surveys below, which cover three broad dimensions: \textit{demography, personality, and morality}.

\begin{figure}[t]
    \centering
    \small
    \includegraphics[width=\linewidth]{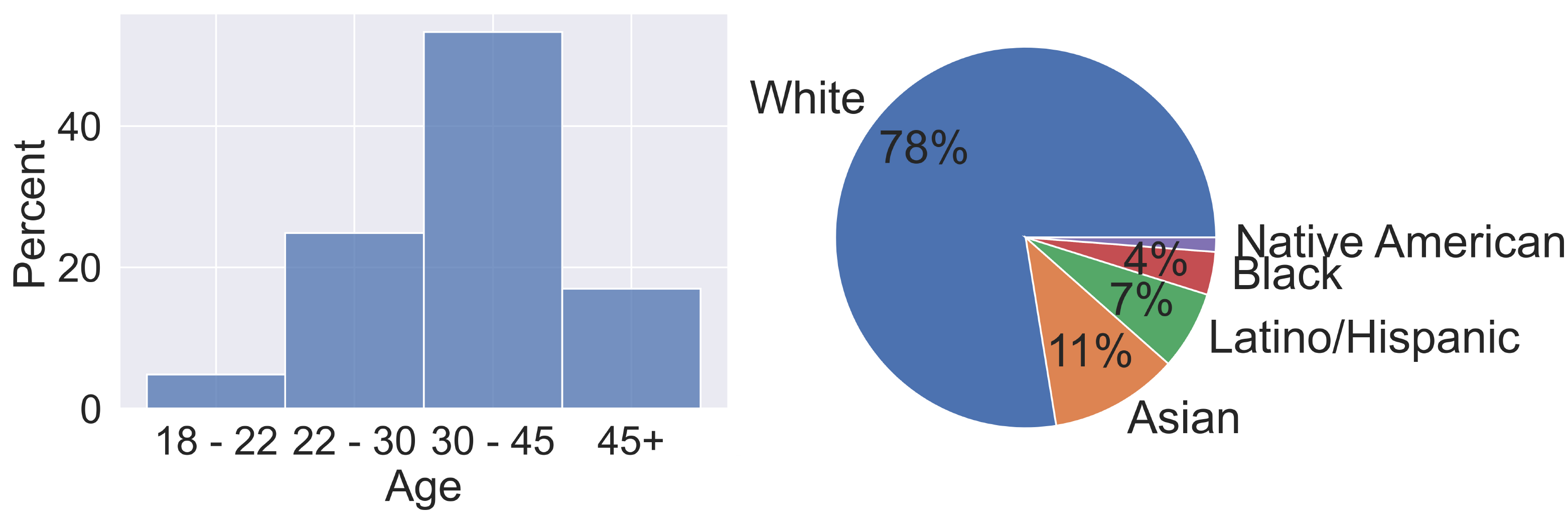}
    \caption{\textbf{Age (left) and Race (right) across our annotators.} Most of our annotators are between 30-45 and are White; however, we still see moderate representation across other racial groups and ages. }
    \label{fig:demo}
\end{figure}

\paragraph{Demographic Data (Figure \ref{fig:demo})} comes from both the annotation UI and in the task's qualifying questionnaires. In the UI, we asked Turkers for their numeric age, their country of origin, and whether English is their native language. These were required features, so we will denote them as \textbf{Demo}$_{\mathbf{Req}}$. In the qualifier, we included an extended demographic survey with \textit{age range, level of education, marital status}, and \textit{native language} (Appendix~\ref{appdx:demographic_questionnaire}), which we will denote as \textbf{Demo}$_{\mathbf{All}}$. We find that our annotator demographics are moderately diverse, mirroring~\citet{moss2020demographic}. Reported gender across annotators are evenly split: 53\% identify as women, 47\% identify as men, and 0\% as other. Additional details are in Figure \ref{fig:demo} and Appendix~\ref{appdx:demographic_questionnaire}.

\begin{figure}[t]
    \centering
    \small
    \includegraphics[width=0.7\linewidth]{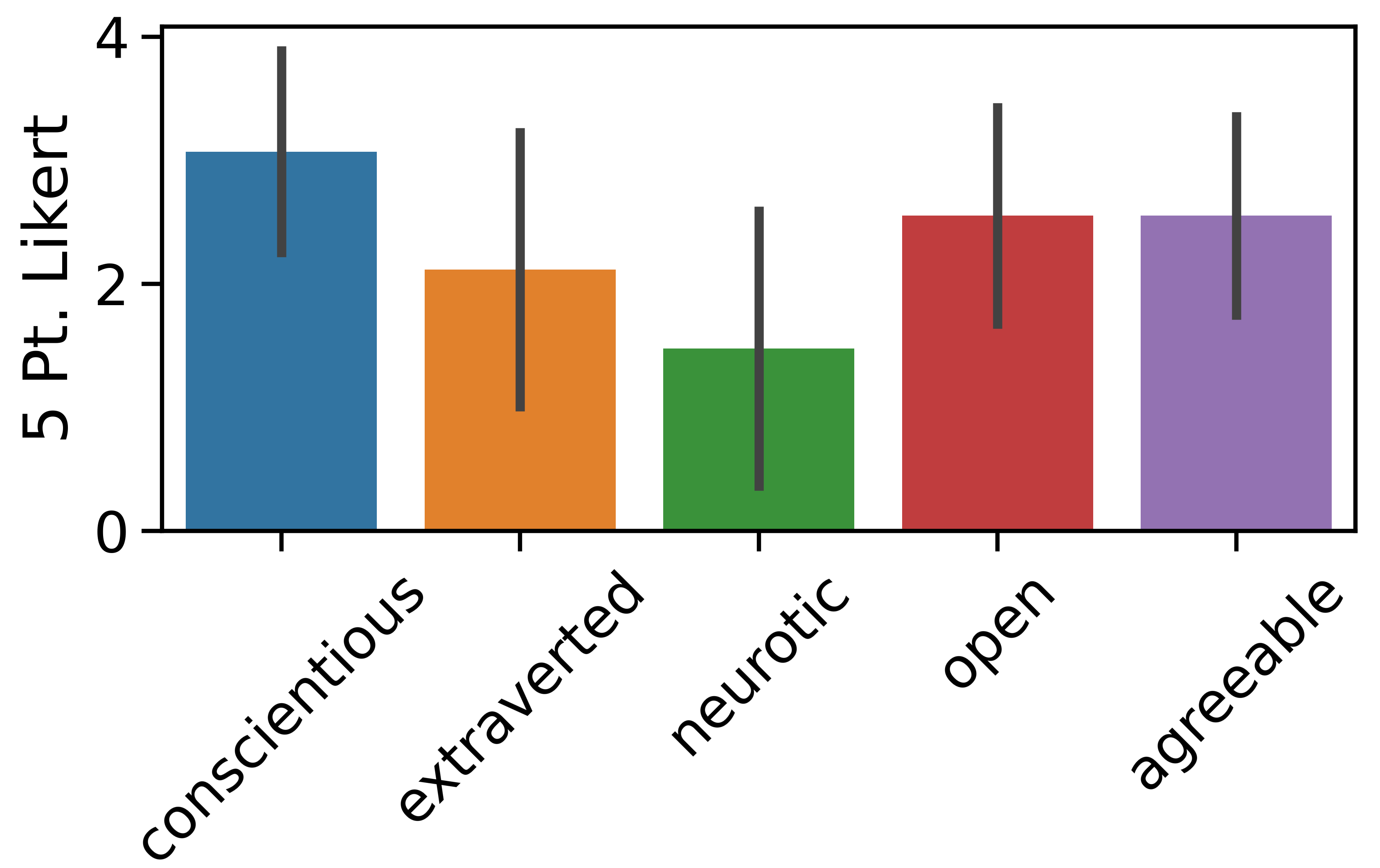}
    \caption{\textbf{Big 5 Personality~\citep{john1991big} results across annotators.} Each personality dimension has a standard deviation $\approx$ 1, indicating a reasonable diversity across our annotator pool. }
    \label{fig:personality}
\end{figure}

\paragraph{Personality (Figure \ref{fig:personality})} surveys also offer insight into interpersonal interactions. We administer the Big 5 Personality Test \citep{john1991big}, measuring a range of personality dimensions on a 5 point Likert Scale. Features include Openness, Conscientiousness, Extraversion, Agreeableness, and Neuroticism. Definitions are in Appendix \ref{appdx:big_5_questionnaire}.

\begin{figure}[t]
    \centering
    \small
    \includegraphics[width=\linewidth]{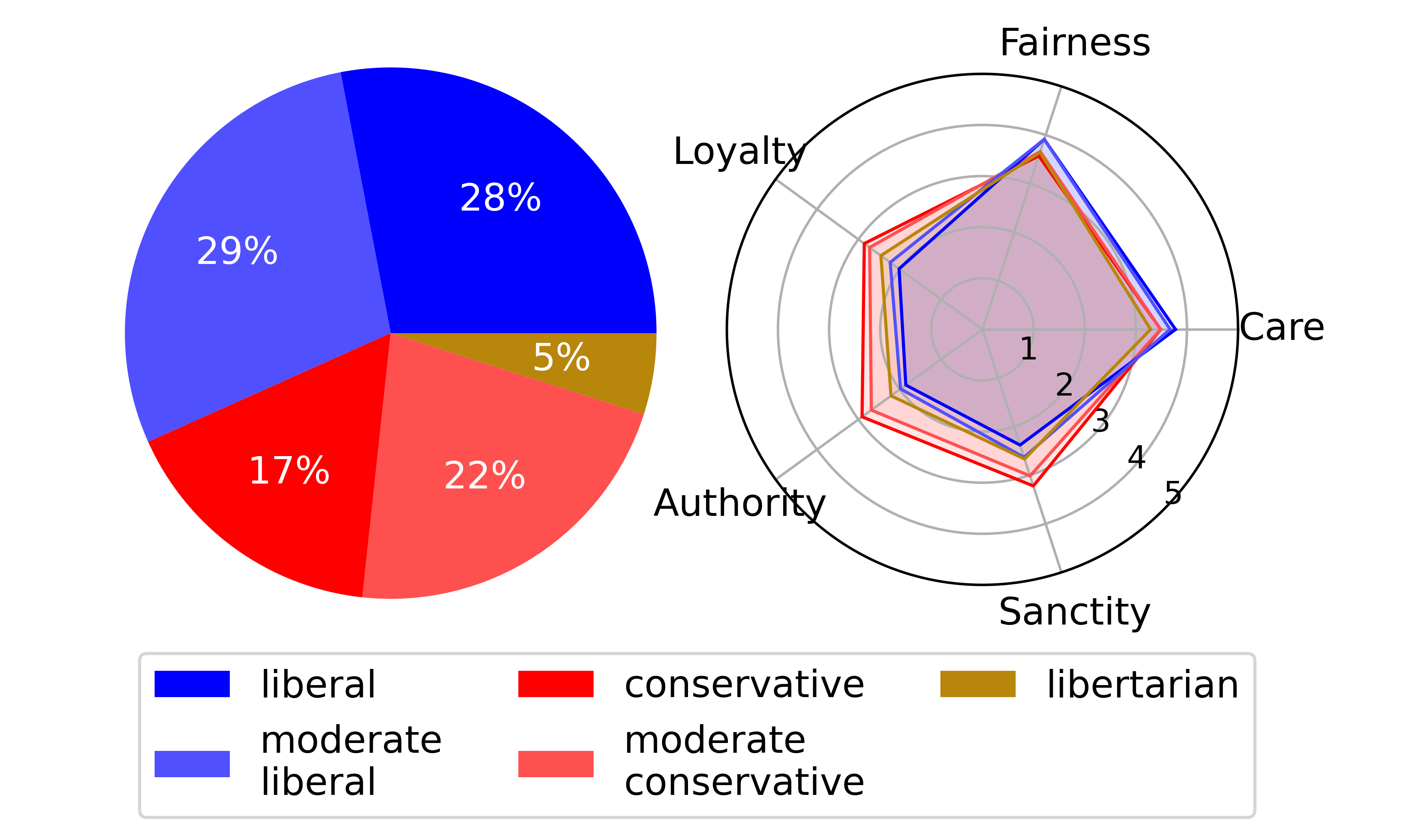}
    \caption{\textbf{Self-reported political leaning (left) and \citet{haidt2007morality}'s Moral Foundations Theory (right) across annotators.} A majority of our workers are liberal (57\%), 39\% are conservative, and the remaining 5\% are libertarian. As observed in \citet{haidt2012righteous}, values like loyalty, authority, and sanctity are higher for conservative leaning annotators, while fairness is higher for liberal annotators ($p < 0.05$, t-test)}
    \label{fig:leanings}
\end{figure}

\paragraph{Moral and Political Leaning (Figure \ref{fig:leanings})} also influences decision making processes. Therefore, we asked annotators to self-report their political leaning (liberal, conservative, libertarian, etc). While political leaning captures broad elements of annotator values, \citet{haidt2007morality}'s widely adopted Moral Foundations Theory (MFT) deconstructs values into individual foundations (Care/Harm, Fairness/Cheating, Loyalty/Betrayal,
Authority/Subversion, and Sanctity/Degradation). Differences in each foundation can stem from cultural variation~\citep{haidt2012righteous}. To record annotator leaning on MFT, we administer an abridged version of the Moral Foundations Questionnaire~\citep{graham2008moral}, which reports each dimension on a 5 point Likert scale (see Appendix~\ref{appdx:mfq_questionnaire}). Later, we refer to all recorded features as \textbf{Morality}.

\begin{table*}[t]
\small
\resizebox{\textwidth}{!}{%
\def\arraystretch{1.15}
\begin{tabular}{lp{3.7cm}p{4.3cm}p{4cm}l}\toprule 
 Agent & Task Description & Input & Output & N  \\ \midrule
 \textsc{Clue Giver} & \textbf{\textit{(1) Target Words}} \newline \underline{Generate}, from the \positive{} $p_i$ words, a subset of targets $t_i$. Targets are used to generate a single clue word.  & $\left\{\textrm{\positive{}}\right\}$ \newline$= \left\{\texttt{BOS}, p_1, p_2,...,p_n, \texttt{EOS}\right\}$ &  $\left\{\textrm{targets}\right\}$ \newline $= \left\{\texttt{BOS}, t_1, t_2,...,t_m, \texttt{EOS}\right\}$ & 7,961 \vspace{2px} \\ 
 \cline{2-5} \\ [-8px] 
 & \textbf{\textit{(2) Generating a Clue}} \newline \underline{Generate} a one word clue $c_1$ that relates selected target words while avoiding \negative{} $a_i$ and neutral $n_i$ words. & $\left\{\textrm{\negative{}}, \textrm{neutral}, \textrm{targets}\right\}$ \newline $ = \left\{\texttt{BOS},\texttt{AVO}, a_1, a_2, \ldots, a_o,\right.$ \newline $\texttt{ }$ $\texttt{ NEU}, \left. n_1, ..., n_n \right.$\newline $\texttt{  }$ $\texttt{ TGT}, \left. t_1, t_2, \ldots, t_m, \texttt{EOS} \right\}$&  $\left\{\textrm{clue}\right\}$ \newline $ = \left\{\texttt{BOS}, c_i, \texttt{EOS}\right\}$ & 7,703 \vspace{2px} \\ \cline{2-5} \\ [-8px] & \textit{\textbf{(3) Framing a Clue}} \newline \underline{Generate} reasoning $r$ that frames a candidate clue word $c_i$ w.r.t. a target $t_i$ word from the set of targets.  & $\left\{\textrm{targets}, \textrm{clue}, \textrm{target} \right\}$ \newline $= \left\{\texttt{BOS}, \texttt{TGTS}, t_1, ..., t_n,\right.$ \newline $\texttt{ }$ $ \texttt{ CLUE},\left. c_i, \texttt{TGT}, t_i, \texttt{EOS} \right\}$ & $\left\{\textrm{rationale}\right\}$\newline $ = \left\{\texttt{BOS}, r, \texttt{EOS} \right\}$ & 9,519 \vspace{2px} \\ \midrule
  \textsc{Guesser} & \textit{\textbf{(4) Selecting Guess Words}} \newline \underline{Generate} a series of guesses $\{g_1, ..., g_m\}$ from the unselected words given a clue $c_i$. & $\left\{\textrm{unselected}, \textrm{clue}\right\}$ \newline $=\left\{\texttt{BOS}, \texttt{UN}, u_1,...,u_n,\right.$ \newline $\texttt{ }$ $\texttt{ CLUE},\left. c_i, \texttt{EOS}\right\}$ & $ \left\{\textrm{guesses}\right\}$\newline$ = \left\{\texttt{BOS}, g_1, g_2,...,g_m, \texttt{EOS}\right\} $ & 7,703 \vspace{2px} \\ \cline{2-5} \\[-8px] & \textit{\textbf{(5) Framing Guesses}} \newline \underline{Generate} reasoning $r$ that frames a guess $g_i$ (from all guesses) w.r.t. clue $c_i$ & $\left\{\textrm{guesses}, \textrm{clue}, \textrm{guess}\right\}$ \newline $= \left\{\texttt{BOS}, \texttt{GUESSES}, g_1, ..., g_n,\right.$ \newline $\texttt{ }$ $ \texttt{ CLUE},\left. c_i, \texttt{GUESS}, g_i, \texttt{EOS} \right\}$  & $\left\{\textrm{rationale}\right\}$ \newline $=\left\{\texttt{BOS}, r, \texttt{EOS} \right\}$ & 9,382\\ \midrule
  \textsc{Both} & \textit{\textbf{Predict Correct Guess}} \newline \underline{Classify} if \textsc{Clue Giver} message (using target, rationale, and clue) is correctly interpreted by the \textsc{Guesser} & $\left\{\textrm{unselected}, \textrm{target}, \textrm{rationale}, \textrm{clue} \right\}$ \newline $= \left\{\texttt{BOS}, \texttt{UN}, g_1, ..., g_n,\right.$ \newline $\texttt{ }$ $ \texttt{ TR},\left. t_i, r_i, \texttt{CLUE}, c_i, \texttt{EOS} \right\}$ & $\{T, F\}$ & 9,519 \\
 \bottomrule
\end{tabular}
}
\caption{\textbf{Tasks associated with a turn in Codenames.} \cluegiver{} starts by selecting information to encode (in the form of a clue), and \guesser{} decodes clues through guesses. In our experiments, we evaluate models with and without sociocultural priors. Task formulation (generation/classification) is \underline{underlined.}}

\label{tab:task_overview}
\end{table*}

\subsection{General Dataset Statistics}
In total, we collect 794 games, with a total of 199 wins and 595 losses.\footnote{Some players went inactive before a game was completed. We only collect games that are reasonably long: greater than the $90^{th}$ percentile of incomplete games, or $\geq$ 7 turns.} Games lasted an average of 9.7 turns, resulting in 7,703 total turns across all games. \cluegiver{} targeted an average of 1.24 words per turn. For all collected games, both players provided Demo$_{\mathbf{Req}}$. For 54\% of games, both players completed all background surveys; for the remaining 46\% of games, at least one player completed all surveys. There were no games with \textit{no} background information.

\section{Tasks and Modeling}
\label{all_tasks}
To investigate the role of sociocultural factors in pragmatic inference, 
we propose a set of tasks (Table \ref{tab:task_overview}) associated with \cluegiver{} (\S \ref{clue_giver_tasks}) and \guesser{} (\S \ref{guesser_tasks}) roles. Concretely,  we formalize each action into a conditional generation problem instead of classification,  
since outputs in \dataset{} are unconstrained: actions and outputs depend on a changing board state.

\subsection{Modeling \textsc{The Clue Giver}}
\label{clue_giver_tasks}

\subsubsection{Selecting Target Words}
\label{gen_target_selection}
To start, \cluegiver{} identifies target word(s) \textbf{(1)} on a board, which are later used to construct a target clue for the inference. Clues will target salient words, where salience is at least partially determined by the speaker's cultural background \citep{wolff2011linguistic}. 
Each set of targets is a subset of the remaining \positive{} words for a given turn ($\mathrm{targets} \subseteq \mathrm{\positive{}}$)---we enforce this restriction in our annotation UI. 

\subsubsection{Giving a Clue}
\label{gen_clue}
After selecting target words, \cluegiver{} must generate a common clue word across the targets \textbf{(2)}. Here, \cluegiver{} must select a prototypical word across the targets. Because cultural background plays a role in inference~\citep{thomas1983cross}, a clue should lie in players' common ground. Furthermore, the clue word should not lead the guesser to pick a \negative{} $n_i$ or \textbf{neutral} $e_i$ word, since these words can end the game or turn (see \S \ref{game_rules}). Therefore, we also include \negative{} and remaining \textbf{neutral} words in our input. 

\subsubsection{Framing the Target Rationales} 
\label{gen_rationale_target}

The relationship between the target and clue word plays a critical role in communication---how information is \textit{framed} with respect to common ground can influence pragmatic success~\citep{crawford2017new}. To this end, we model \cluegiver{}'s framing of the rationale $r$ for a specific target word $t$ \textbf{(3)}, connecting the target $t$ to the clue (c.f., \S \ref{gameplay_data}). Because the framing is constructed in relation to every target word (if multiple are provided), we also encode all targets in the input. 

\begin{table}[t]
\small
\centering
\def\arraystretch{1.15}
\begin{tabular}{ll|l|l}\toprule

 Priors & Model & {Target R-1} & {Guess R-1}   \\ \midrule
 \multirow{4}{*}{\shortstack{\textbf{No Priors}}} &Random& 0.60 & 0.65 \\
&$k$-NN fastText& N/A & 58.04  \\ 
&T5& 32.57 & 64.96 \\
 &BART& 31.82 & 63.30  \\ \midrule
 \multicolumn{4}{c}{$\downarrow$ \textbf{\textit{With} Sociocultural Priors}} \\
 \midrule
Demo$_{\mathbf{Req}}$ &T5& 32.71 & \hspace{-.6em} \best{67.25} \\
 &BART& 29.45 & 65.18  \\ \midrule
Demo$_{\mathbf{All}}$ &T5& 33.14 & 65.24 \\
 &BART& 32.27 & 66.02  \\ \midrule
Personality &T5& 33.61 & 65.56 \\
 &BART& 28.55 & 63.14  \\ \midrule
Morality &T5& \hspace{-.6em} \best{34.58} & 64.60 \\
 &BART& 31.32 & 65.09  \\ \midrule
All &T5& 33.38 & 66.31 \\
 &BART& 30.17 & 64.78  \\
\bottomrule 
\end{tabular}

\caption{\small{\textbf{Target (\S \ref{gen_target_selection}) \& Guess (\S \ref{gen_guess_selection}) Selection Generation Results}. We report only R-1 scores, since tasks must contain exact single-word matches to reference labels. Target Selection is maximized when using Morality priors, while Guess Selection is maximized by using only Demo$_{\mathbf{Req}}$.}}
\label{tab:guess_tgt_selection_results}
\end{table}

\begin{table}[t]
\small
\centering
\def\arraystretch{1.15}
\begin{tabular}{ll|ll}\toprule
 Priors & Model & {Clue R-1} & {fastText $cos$}   \\ \midrule
\multirow{4}{*}{\shortstack{\textbf{No Priors}}} & Random & 0.08 & 5.76 \\
&$k$-1 fastText& 0.00 & 10.33  \\
 &T5& 23.86 & 40.38 \\
 &BART& 23.00 & 40.97  \\ \midrule
 \multicolumn{4}{c}{$\downarrow$ \textbf{\textit{With} Sociocultural Priors}} \\
 \midrule
Demo$_{\mathbf{Req}}$ &T5& 25.47 & 42.91 \\
 &BART& 20.64 & 38.91 \\ \midrule
Demo$_{\mathbf{All}}$ &T5& 25.74 & 42.07 \\
 &BART& 21.45 & 39.45 \\ \midrule
Personality &T5& 24.13 & 41.00 \\
 &BART& 23.32 & 41.49 \\ \midrule
Morality &T5& \hspace{-.6em} \best{26.54} & 43.31 \\
 &BART& 23.59 & 41.39  \\ \midrule
All &T5& 26.27 & \hspace{-.6em} \best{44.03} \\
 &BART& 24.40 & 41.60 \\
\bottomrule 
\end{tabular}

\caption{\small{\textbf{Clue Generation Results (\S \ref{gen_clue}) } We report R-1 scores and fastText $cos$ similarities between the reference and generation, since outputs must be semantically close to or exactly match the reference labels.} We find that \textbf{Morality} and \textbf{All} maximize performance over our metrics. }
\label{tab:gen_clue_results}
\end{table}

\subsection{Modeling \textsc{The Guesser}}
\label{guesser_tasks}

\subsubsection{Selected Guesses} 
\label{gen_guess_selection}
With the clue word, the \guesser{} pragmatically infers \cluegiver{}'s targets, selecting a sequence of corresponding guesses \textbf{(4)}. For this task, we model the sequence of all selected guesses, regardless of correctness. We input all \textit{unselected}\footnote{Note that \positive{}/\negative{}/\textbf{neutral} words differ across players. A \positive{} word for one player can be \negative{} for another; game states are asymmetric. A clue from \cluegiver{} may also target a \positive{} word for the \guesser{}. As long as one does not guess a \negative{} word from the \textit{opposing} player, the game continues. See \S \ref{game_rules}.} words at the start of each turn for \guesser{}, along with the provided clue. Like with Target Word Selection, guesses must be a subset of the unselected words ($\mathrm{guesses} \subseteq \mathrm{unselected}$); we enforce this during annotation.

\begin{table*}[ht!]
\small
\resizebox{\textwidth}{!}{%
\def\arraystretch{1.15}
\begin{tabular}{ll|lllll|lllll}\toprule
& & \multicolumn{5}{c|}{\textbf{Target Framing}} & \multicolumn{5}{c}{\textbf{Guess Framing}} \\
 Priors & Model & {R-1} & {R-2} & {R-L} & {BLEU} & {BScore} &  {R-1} & {R-2} & {R-L} & {BLEU} & {BScore} \\ \midrule
 \multirow{4}{*}{\shortstack{\textbf{No Priors}}} &Random& 14.08 & 3.80 &  13.88 & 3.46 & 86.88 & 8.31 & 1.01 & 8.07 & 0.80 & 85.88  \\
 &SBERT& 53.14 & 23.10 & 49.13 & 20.04 & 92.24 & 40.49 & 10.82 & 33.57 & 10.53 & 89.31 \\  
  &T5& 69.22 & 36.82 & 64.13 & 34.11 & 94.52 & 54.67 & 19.65 & 47.22 & 17.40 & 91.25  \\
 &BART& 66.20 & 31.85 & 59.84 & 30.09 & 93.72 & 52.36 & 17.27 & 44.49 & 14.72 & 90.85  \\ \midrule
 \multicolumn{12}{c}{$\downarrow$ \textbf{\textit{With} Sociocultural Priors}} \\
 \midrule
Demo$_{\mathbf{Req}}$ &T5& 70.15 & 37.86 & 64.81 & 35.05 & 94.61 & 57.26 & 23.19 & 48.32 & 23.31 & 91.63 \\
 &BART& 67.16 & 34.52 & 60.97 & 31.47 & 94.00 & 54.55 & 19.11 & 45.69 & 17.62 & 90.95  \\ \midrule
Demo$_{\mathbf{All}}$ &T5& \hspace{-0.6em} \best{70.40} & 38.14 & 64.98 & 35.07 & 94.60 & 57.22 & 23.14 & 48.36 & 21.05 & 91.59 \\
 &BART& 66.14 & 32.21 & 59.72 & 31.36 & 93.88 & 52.43 & 16.51 & 43.52 & 13.23 & 90.78 \\ \midrule
Personality &T5& 69.68 & 38.31 & 64.74 & \hspace{-0.6em} \best{35.27} & 94.47 & 57.41 & 23.08 & 48.72 & 21.37 & 91.61  \\
 &BART& 67.12 & 34.36 & 61.34 & 32.10 & 93.88 & 52.89 & 18.85 & 45.07 & 15.55 & 90.92 \\ \midrule
Morality &T5& 69.82 & 37.96 & 64.35 & 34.53 & 94.63 & \hspace{-0.6em} \best{58.06} & \hspace{-0.6em} \best{23.67} & \hspace{-0.6em} \best{48.85} & \hspace{-0.6em} \best{22.62} & \hspace{-0.6em} \best{91.76} \\
 &BART& 67.78 & 34.47 & 61.49 & 32.25 & 94.25 & 53.46 & 18.49 & 45.73 & 14.95 & 90.93 \\ \midrule
 All &T5& 70.39 & \hspace{-0.6em} \best{38.27} & \hspace{-0.6em} \best{65.49} & 34.01 & \hspace{-0.6em} \best{94.66} & 57.64 & 23.13 & 48.79 & 22.22 & 91.68 \\
 &BART& 67.66 & 34.45 & 62.28 & 31.59 & 93.95 & 52.12 & 18.13 & 44.51 & 15.96 & 90.92 \\
\bottomrule 
\end{tabular}
}
\caption{\small{\textbf{Framing Generation Results} for Target (\S \ref{gen_rationale_target}) and Guess (\S \ref{gen_rationale_guess}) words. } We find that the best models with sociocultural priors \textbf{universally} outperform their baseline counterparts. For Target Rationale Generation, jointly modeling all features yields highest improvements; Guess Rationale generation sees improvements when using Morality priors. Guess Rationale Performance sees higher relative/absolute improvement from baselines compared to Target Rationale Generation. }

\label{tab:generation_results}
\end{table*}

\begin{table}[]
    \centering
    \small
    \begin{tabular}{l|llll}
    \toprule
    Priors & Random & BERT & RoBERTa & XLNet \\
    \midrule
    \textbf{None} & 0.50 & 0.57 & 0.57 & 0.57 \\ \midrule
 \multicolumn{5}{c}{$\downarrow$ \textbf{\textit{With} Sociocultural Priors}} \\
 \midrule
    Demo$_{\mathbf{Req}}$ & -- & 0.52 & 0.55 & 0.52 \\
    Demo$_{\mathbf{All}}$ & -- & \textbf{0.59} & 0.63 & 0.62 \vspace{-.2em} \\ 
    Personality & -- & 0.57 & \hspace{-.6em} \best{\textbf{0.67}} & \textbf{0.64} \\
    Morality & -- & 0.57 & 0.64 & 0.61 \\
    All & -- & 0.57 & 0.65 & 0.63 \\
    \bottomrule
    \end{tabular}
    \caption{\small Macro F-1 scores for \textbf{Predicting Pragmatic Success (\S \ref{cla_correct_guess})}: models must predict if a guesser will guess correctly given the target word, target rationale, and clue. We use base variants of all models and experiment with ablations across different background characteristics. }
    \label{tab:correct_guess_confidence}
\end{table}

\subsubsection{Framing Guess Choice} 
\label{gen_rationale_guess}
Finally, \guesser{} also provides framing rationale for their respective guesses, framing clues with respect to their guess \textbf{(5)}.

\subsection{Predicting Pragmatic Success}
\label{cla_correct_guess}
So far, our tasks focus on \textit{replicating} elements of a game turn: the Selected Guesses task (\S\ref{gen_guess_selection}), for example, models both incorrect and correct guesses. However, we also wish to understand if an entire turn sequence results in a \textbf{successful} inference; differences in cross-cultural inferences can result in pragmatic failures~\citep{thomas1983cross}. We formulate this as binary classification. 

Importantly, we only consider a guess correct if it is \textit{intentional}. A guess is intentional \textit{if and only if} the clue giver listed it as a target. If \guesser{} selects a \positive{} word that is \textit{not} a target word, we count it as ``incorrect.'' Like with guess generation, we encode unselected words in the input. Because we are not predicting the guess itself, we include target and rationale from \cluegiver{}. 

\subsection{Augmenting with Sociocultural Priors}
\textbf{We hypothesize that players' backgrounds influence Codenames gameplay.} To this end, we encode background player information for each task. For each dimension described in \S\ref{demographic_info}, we encode an attribute/answer pair (e.g. \texttt{age: 22}) for each survey question. Then, we prepend all attributes to the encoded strings for each outlined task (\S\ref{all_tasks}), using a unique token to delimit attributes for \cluegiver{} and \guesser{}.
$$
\begin{aligned}
\boldsymbol{in_{socio}} = &\left\{\texttt{BOS}, \texttt{GIVER}, \textrm{Clue Giver}_{\mathrm{Attr:A}}, \right. \\ &  \texttt{ GUESSER}, \left. \textrm{Guesser}_{\mathrm{Attr:A}}\right\} + \boldsymbol{in}
\end{aligned}
$$
If a player did not respond to a specific attribute, we replace the attribute/answer pair with \texttt{None}. From our sociocultural priors (\S \ref{demographic_info}), we have 5 ablations: Demo$_{\mathbf{Req}}$, Demo$_{\mathbf{All}}$, Personality, Morality, and All (concatenating and modeling all ablations). We additionally use \textit{no} priors as a baseline, using $\boldsymbol{in}$ instead of $\boldsymbol{in_{socio}}$ to test our hypothesis. 

\section{Experiment Setup}
\paragraph{Baselines and Dataset Splits} 
For generation baselines, we use two Seq2Seq models: T5~\citep{raffel2020exploring} and BART~\citep{lewis2019bart}. We optimize the associated language modeling objective across our tasks. 
Additionally, we experiment with two retrieval baselines for all generation tasks: (1) randomly selecting a generation from the train set and (2) selecting the nearest $k$-N inputs using pretrained SentenceBERT~\citep{reimers-2020-multilingual-sentence-bert} or fastText~\citep{bojanowski2016enriching}.  Retrieval baselines yield insight into how well off-the-shelf pretrained models capture sociocultural diversity. 
For classification, we experiment with BERT \citep{devlin-etal-2019-bert}, RoBERTa \citep{liu2019roberta}, and XLNet \citep{yang2019xlnet}. Models are base variants, and results are averaged over 5 runs. 

For each task, we split \textit{clue givers} into 80-10-10 train/val/test, since all tasks depend on initial clue giver choices. Importantly, \textbf{a single clue giver's data is not distributed across splits}, since clue givers may reuse clues/strategies. 

\paragraph{Evaluation Metrics} We use a range of metrics to generation tasks. Rationale generation tasks (Target \S \ref{gen_rationale_target} \& Guess \S \ref{gen_rationale_guess}) output entire sentences; therefore, we report F-1 scores from ROUGE-(1, 2, L)~\citep{lin2004rouge}, BLEU~\citep{papineni2002bleu}, and BERTScore~\citep{bert-score}. For tasks that generate a single or set of words where order does not matter, (Guess Selection \S \ref{gen_guess_selection}; Clue Generation \S \ref{gen_clue}), we report only ROUGE-1 and averaged word vector (fastText) cosine similarity.

\section{Generation Results \& Discussion}
\label{results}
Including cultural priors improves modeling performance across \textbf{all} tasks. For generation problems, T5 generally outperforms BART, and our retrieval baselines lag behind more complex models. Finally, we conduct \qual{} \textit{a qualitative analysis} of 20 random samples from each task.

\paragraph{Picking Targets and Guesses} From our results (Table \ref{tab:guess_tgt_selection_results}), we find that selecting guesses is an easier modeling task than picking target words, likely because the input for selecting a guess contains the clue word. Intuitively, selecting target words is more arbitrary than selecting a guess from a clue---especially since our generation task does not enforce guess correctness. Our models reflect this observation. Guess Selection has R-1 scores that are, on average, twice as good as Target Word Selection (Target $34$ vs. Guess $66$). Furthermore, Guess Selection only requires demographics (Demo$_{\mathbf{Req}}$) to maximize performance, unlike \textbf{Morality} for Target Words. Regardless, both tasks see R-1 increase by $\approx2$ points over no prior baselines. 

\qual{} Looking at model outputs between the \textbf{None} and \textbf{Morality}, we observe that models generate words like \textit{Well}/\textit{Grace} instead of \textit{Death}/\textit{Poison} and vice versa, depending on player background. 

\paragraph{Generating a Clue for Targets} 

Moving to our clue generation models, we again find that including sociocultural priors improves model performance (Table \ref{tab:gen_clue_results}). Highest R-1 scores (26.54) occur when using \textbf{Morality} as a prior, resulting in a $\approx2$ pt. R-1 and $4$ pt. cos-similarity increase when compared to a no prior baseline. We also suspect that selecting target words and generating a hint are interrelated processes: annotators are likely thinking about clues/targets in parallel. Therefore, the same Morality prior results in maximized performance.

\qual{} While there are themes related to Morality in clue differences for a \underline{target} word (\underline{accident} $\rightarrow$ death vs. lucifer; or \underline{fair} $\rightarrow$ equal vs. good), we also find that generations are \textit{more specific} given sociocultural priors. Consider these generated \underline{target} $\rightarrow$ clue pairs \cmark{} with and \xmark{} without priors: 
\begin{itemize}
    \setlength\itemsep{0em}
    \item \underline{match} $\rightarrow$ \xmark{} game \cmark{} cricket
    \item \underline{bond} $\rightarrow$ \xmark{} connection \cmark{} james
    \item \underline{undertaker} $\rightarrow$ \xmark{} funeral \cmark{} wrestler
\end{itemize}

Each \cmark{} example generates a clue that relies on shared cultural background: specifically, knowing that cricket is a sport; that James Bond is a popular character; and that the Undertaker is a wrestler. More details can be found in Appendix \ref{appdx:generations}, Table \ref{tab:my_label}.

\paragraph{Clue Generation Errors Across Sociocultural Subtypes} Despite jointly modeling cross-cultural information, our performance is far from perfect. Generating successful clues is a core element of Codenames; however, our exact match accuracy on clue generation is only $\approx$ 26\%. To understand errors, we sample 100 generated clues from the Clue Generation Task, and identify errors and differences between (socioculturally) generated clues and the ground truth label. 

For 43 samples, we notice that sociocultural priors have \textit{no effect} on clue generation; the output is identical to the \textit{no prior} model for the given target word. In these instances, we suspect that our models fail to exploit common ground between a giver/guesser, yielding the same clue as without sociocultural priors. Upon further analysis, we observe that these errors occur frequently (37 samples) when \textit{both} the clue giver and guesser are white or from North America. Because these demographics are already over-represented in our dataset, we suspect that the model simply ignores over-informative sociocultural priors.

Errors also occur because clues are over (20 instances, e.g. ``guevera'' instead of ``overthrow'') or underspecified (13 instances, e.g. ``supernatural'' instead of ``monster'') compared to the gold clue. In 21/33 of these instances, there is a demographic mismatch between the clue-giver and guesser: the clue-giver and guesser do not share race/country demographics. In contrast to having no effect, we suspect that models mispredict the common ground between guesser/giver. We also judge 18 generation errors to be of similar specificity to the target word---prefixes/suffixes of the gold label---or completely unrelated to the gold clue (6 instances).
 
\paragraph{Rationalizing Targets and Guesses} Beyond generating target words and guesses, we ask models to explain how a \underline{target} or guess is related to a clue word (e.g. James Bond is a movie character). Again, we find that providing contextual priors improves performance (Table \ref{tab:generation_results}). For Target Rationale Generation, models see maximized performance when \textbf{all} priors are included, while Guess Rationale generation sees improvements for \textbf{Morality.}

\qual{} Like with Clue Generation, we find that improvements in Guess Rationale are from increased specificity (e.g. ``actors are cast'' $\rightarrow$ ``actors are part of a cast''; ``money is center'' $\rightarrow$ ``money is the center of everything''). While qualitative differences are clear for Guess Rationale, Target Rationale results are more subtle: improvements stem from minor variations in the type of framing ("a kind of" vs. "a type of") used by the annotator. Additional generations can be found in Appendix \ref{appdx:generations}, Table \ref{tab:clue_framing_additional}.

\paragraph{Classifying Pragmatic Failure} We find that classification performance across each architecture is maximized when using sociocultural priors during training (Table \ref{tab:correct_guess_confidence}). While BERT sees reduced improvement (an increase of only +0.02 F-1 over a no-prior baseline), XLNet and RoBERTa see maximum increases of +0.07 and +0.10 respectively. Both XLNet and RoBERTa see these improvements across the same \textbf{Personality} setting. Sociocultural priors improve performance across mirroring \textit{and} evaluating pragmatic inference.

\paragraph{A Word on Word Vector Baselines} Surprisingly, retrieving nearest words using a word vector approach (fastText) performs poorly for both Clue and Guess Generation (Tables \ref{tab:guess_tgt_selection_results} \& \ref{tab:gen_clue_results}). We suspect that pretrained vectors fail to capture sociocultural inference in word association tasks.

\section{Conclusion}
Language is grounded in rich sociocultural context. To underscore this context, we propose a setting that captures the diversity of pragmatic inference \textit{across} sociocultural backgrounds. With our Codenames Duet dataset (7K turns across 156 players), we operationalize cross-cultural pragmatic inference. Across our experiments, we detail improvements in mirroring/evaluating inferences when using sociocultural priors. Our work highlights how integrating these priors can align models toward more socially relevant behavior. 

\section{Limitations}

\paragraph{Cross-Cultural Inference Beyond Codenames} Our work explores sociocultural pragmatic inference in a very limited setting, using a core vocabulary of just 100 words. Despite this limitation, we find significant diversity in our dataset; furthermore, our models successfully capture these diverse inferences. While a limitation of our work is its focus on a single setting, we expect domains outside of Codenames to see similar variance. Understanding and highlighting miscommunication in dialog---due to culture-dependent misinterpretation---is one such extension. These domains are likely much nosier than Codenames; we urge future work to further investigate them. 

\paragraph{Spurious Correlations across Sociocultural Factors} Across all tasks but one (Target Rationale Generation \S \ref{gen_rationale_target}), jointly modeling all sociocultural priors does not result in the highest performing model. Because our sociocultural factors already correlate with each other (\S \ref{demographic_info}), we suspect that modeling all features may be redundant, adding spurious correlations and resulting in overfitting. Improved modeling methodology and careful regularization may address these issues; we leave these experiments for future work.  

\paragraph{Bigger Models and Task Specific Modeling} Currently, we evaluate small Seq2Seq models due to computational constraints; however, evaluation of 0-shot and few-shot performance on larger language models (e.g. GPT-3) is necessary. Given the changing state of the Codenames board---along with evidence that LLMs struggle with theory-of-mind-esque perspective taking~\citep{sap2022neural}---our dataset can serve as a challenging benchmark for sociocultural understanding. However, successfully encoding game state into prompts for LLMs may require experimentation. 

Finally, our current task formulation and modeling setup are straightforward: we simply encode all information \textit{in-context} and do not assume recursive reasoning like in RSA~\citep{goodman2016pragmatic}.  Future work can explore these directions. 

\paragraph{Human Evaluations} Our evaluation is limited to automatic metrics and qualitative analysis. Evaluating cross cultural generation \textit{depends} on the evaluator's own culture. Each generation depends on the player's sociocultural background; finding evaluators who match the player may be prohibitive.  

\section{Ethics}
Broadly, our work models user background to determine the choices they make. While we focus on a fairly harmless setting (Codenames), our operationalization can be used in harmful ways (e.g. tracking and modeling user behavior without consent). Future work that uses sociocultural information should only be applied to settings where there is no foreseeable harm to end-users.

Furthermore, learning sociocultural associations can introduce positive and negative stereotypes; documenting and reducing harmful stereotypes is an important avenue for future work. Finally, we emphasize that our work is not evidence for \textit{linguistic determinism}: sociocultural variation in language can influence but not \textbf{determine} thought. 

\section*{Acknowledgements}  We are thankful to the members of SALT Lab for their helpful feedback on the draft. We are also thankful for the helpful feedback from Jing Huang and Rishi Bommasani. Caleb Ziems is supported by the NSF Graduate Research Fellowship under Grant No. DGE-2039655. This research was supported, in part, by MURI-ONR-N00014-20-S-F003 on Persuasion, Identity, and Morality in Social-Cyber Environments, 
as well as a DARPA grant HR00112290103/HR0011260656. 

\bibliographystyle{acl_natbib}
\bibliography{ref}

\appendix
\section{Finalized Codenames Word List}
\label{word_list}
We sample from the following list of 100 words: \textit{luck, grace, soul, fair, life, pass, revolution, change, charge, degree, force, code, genius, compound, time, wake, plot, draft, ghost, play, part, spell, well, point, link, mass, disease, sub, state, alien, space, mine, ray, millionaire, agent, bond, unicorn, figure, war, cycle, boom, sound, trip, centaur, death, club, crash, angel, cold, center, spring, round, date, press, cast, day, row, wind, fighter, embassy, beat, leprechaun, comic, pitch, mount, march, fall, undertaker, green, switch, strike, king, superhero, capital, slip, lead, check, lap, mammoth, air, match, spy, roulette, contract, witch, stock, light, drop, spot, novel, vacuum, cover, scientist, tag, conductor, field, racket, poison, ninja, opera.}

\section{Reformatting Rationales using GPT-3}
\label{gpt3reformat}
Some annotators wrote verbose rationales (\textit{I think fall happens after you slip}), while other annotators were more succinct (\textit{fall after slip}). To prevent models from learning grammar variation across annotators, we normalize our text using GPT-3. We use the following prompt, using hand-written few-shot examples. Some of the examples are unchanged---we include them in the prompt to demonstrate positive examples to the model.

\vspace{.5em}

\noindent
\texttt{Normalize the text, removing determiners like ``the'' and ``a'' at the start of a sentence, along with any pronouns. Correct spelling and grammar mistakes. If possible, the final text should be formatted with the clue first and the target last or the target first and the clue last. \\ \\ 
clue: ``sub''\\
target: ``sandwich''\\
text: ``you can make a sub, which is a type of sanwich''\\
output: ``sub is a type of sandwich''\\ \\
clue: "die"\\
target: "cliff"\\
text: "you may die if you fall off a cliff"\\
output: "die if fall off a cliff"\\ \\
clue: "explosion"\\
target: "boom"\\
text: "it makes sound"\\
output: "explosion makes boom"\\ \\
clue: "superman"\\
target: "superhero"\\
text: "most famous superhero"\\
output: "superman is most famous superhero"\\ \\
clue: "night" \\
target: "club" \\
text: "i love night club" \\
output: "night club is a kind of club" \\ \\
clue: "horn" \\
target: "air" \\
text: "an air horn is a type of horn" \\
output: "air horn is a type of horn" \\ \\
clue: "ivy" \\
target: "poison" \\
text: "poison ivy is a well known plant" \\
output: "poison ivy is a well known plant" \\ \\
clue: "month" \\
target: "march" \\
text: "march is a month" \\
output: "march is a month" \\ \\
clue: "\{clue\}" \\
target: "\{target\}" \\
text: "\{text\}" \\
output: "
}

\section{Example Generations}
\label{appdx:generations}
Here, we include example generations for a subset of our tasks, illustrating the influence of sociocultural factors on generated Codenames gameplay. 

\subsection{Clue Generation}
Below, we highlight more clues generated with/without sociocultural priors. Note how some of the without generations are euro-centric: space $\rightarrow$ nasa, \{revolution, king\} $\rightarrow$ war; adding priors creates more specific clues. However, this isn't always true: target words \{pass, check\} $\rightarrow$ leads to poker instead of overtake when conditioned on priors. We suspect that the average player in our pool is not aware of how \{pass, check\} are associated with poker, resulting in a more generic generation. 

\begin{table}[h]
    \centering
    \begin{tabular}{c|ccc}
        \toprule
        Target & Without & With & Gold \\
        \midrule
        revolution, king & war & guevara & overthrow \\
        check & mate & inspect & examine \\
        space & nasa & galaxy & universe \\
        compound & wall & house & together \\
        pass, check & overtake & poker & go \\
        \bottomrule
    \end{tabular}
    \caption{Clue generations with/without sociocultural priors, given target words on the board}
    \label{tab:my_label}
\end{table}

\subsection{Clue Framing}
Additional generations can be found in Table \ref{tab:clue_framing_additional}. Again, we observe that adding sociocultural priors increases relation specificity.
\begin{table*}
    \centering
    \begin{tabular}{cc|p{3.5cm}p{3.5cm}p{3.5cm}}
        \toprule
        Target & Clue & Without & With & Gold \\
        \midrule
        explode & boom & explode causes boom & bomb explodes \textbf{with a} boom & explosions make a boom sound\\
        horse & unicorn & a unicorn is a horse & unicorn is \textbf{a type of} horse & unicorns are similar to horses\\
        racket & tennis & tennis has racket & a racket \textbf{is used in} tennis & tennis uses a racket\\
        day & month & day is month & month \textbf{has many} days & 30 days in a month \\
        \bottomrule
    \end{tabular}
    \caption{Example Rationales for Clues, with/without background priors. With priors, we observe that rationales become more specific, mentioning explicit relations between the target and clue.}
    \label{tab:clue_framing_additional}
\end{table*}

\section{Annotation Task Details}

\subsection{Qualification Test}
\label{appdx:qualification_test}
To qualify for the HIT, workers were required to complete a consent form detailing dataset collection and release; and were expected to watch an instructional video outlining game rules.

Then they had to pass the following qualifying test, answering at least 6 out of 7 questions correctly.

\begin{enumerate}
    \item \textbf{True or False:} "angry dog" is an example of a clue you could give. [\textit{Answer}: \textbf{False}]
    \item \textbf{True or False:} you and your partner have different lists of black (assassin) words. [\textit{Answer}: \textbf{True}]
    \item \textbf{True or False:} it is possible to skip a turn without guessing. [\textit{Answer}: \textbf{False}]
    \item \textbf{True or False:} the tan ``down'' arrow indicates that you guessed the word wrong, while the tan ``up'' arrow indicates that your partner guessed it wrong. [\textit{Answer}: \textbf{True}]
    \item \textbf{Multiple Choice:} Which of the following kinds of phrases does not follow from our list of target rationales types? [\textit{Answer}: \textbf{(b)}]
    \begin{enumerate}
        \item ``a computer has a mouse''
        \item ``a doctor is smart''
        \item ``a dog is a kind of animal''
        \item ``a disease causes people to be sick''
    \end{enumerate}
    \item \textbf{Multiple Choice:} How many guesses do you get (assuming there are still more words left to guess) [\textit{Answer}: \textbf{(d)}]
    \begin{enumerate}
        \item you get three guesses each turn
        \item the number of guesses you get is the same as the number of target words your partner's clue
        \item as long as you keep picking green words, you can keep guessing, up to the number of target words in your partner's clue
        \item as long as you keep picking green words, you can keep guessing without any limit, even if you guess more than the number of target words in your partner's clue
    \end{enumerate}
    \item \textbf{Multiple Choice:} During the 8th timer token in the video, it looked like my grid froze and I couldn't make any more guesses. Why did this happen? [\textit{Answer}: \textbf{(b)}]
    \begin{enumerate}
        \item I guessed an assassin word
        \item I already guessed all my partner's words correctly
        \item I clicked the ``end game'' button
        \item My partner left the game
    \end{enumerate}
\end{enumerate}

\subsection{Demographic, Personality, and Moral Questionnaires}
\label{appdx:questionnaires}
Before starting any HITs, workers also had to complete three standardized surveys about their moral foundations, personality, and demographic information. The survey questions and worker statistics are given as follows.

\subsubsection{Worker Demographics}
\label{appdx:demographic_questionnaire}
\paragraph{Questionnaire.} Please answer these 8 questions about yourself.
\begin{enumerate}
    \item With what gender do you identify? \{\textit{Woman, Man, Transgender, Non-binary / non-conforming, Other}\}
    \item What is your age? \{\textit{0-17 years old, 18-22 years old, 22-30 years old, 30-45 years old, 45+}\}
    \item Which best describes your race or ethnicity? \{\textit{African-American/Black, Asian, Latino or Hispanic, Native American, Native Hawaiian or Pacific Islander, White / Caucasian}\}
    \item In which continent are you located? \{\textit{North America, Central / South America, Europe, Africa, Asia, Australia}\}
    \item What is your highest level of education? \{\textit{Some High School / No Diploma, High School Diploma, Associate's Degree / Trade School, Master's Degree, Doctorate Degree}\}
    \item What is your marital status? \{\textit{Single and never married, Married or in a domestic partnership, Widowed, Divorced, Separated}\}
    \item Which of the following would you consider your native language \{\textit{English, Arabic, French, Mandarin, Spanish, Other}\}
    \item If applicable, please specify your religion \{\textit{Buddhism, Catholicism/Christianity, Hinduism, Islam, Judaism, Other}\}
\end{enumerate}

\paragraph{Results.} Of the 153 unique players, 124 are from the U.S, 12 are from India, 8 are from Brazil, 3 from the U.K, 2 from Canada, and the rest are single players from the following 7 countries: Indonesia, Costa Rica, France, South Africa, Germany, and Portugal.

\subsubsection{Worker Personality} 
\label{appdx:big_5_questionnaire}
\paragraph{Big 5 Personality Questionnaire.}
Please answer these 10 questions about yourself on the following scale: [-2] Strongly Disagree; [-1] Disagree; [0] Neutral; [1] Agree; [2] Strongly Agree.
\begin{enumerate}
    \item I see myself as someone who does a thorough job.
    \item I see myself as someone who is reserved.
    \item I see myself as someone who is outgoing, sociable.
    \item I see myself as someone who gets nervous easily.
    \item I see myself as someone who has few artistic interests.
    \item I see myself as someone who is relaxed, handles stress well.
    \item I see myself as someone who tends to find fault with others.
    \item I see myself as someone who is generally trusting.
    \item I see myself as someone who tends to be lazy.
    \item I see myself as someone who has an active imagination.
\end{enumerate}

\subsubsection{Moral Foundations And Political Leaning.}
\label{appdx:mfq_questionnaire}

\paragraph{Moral Foundations Theory.} Following \citet{haidt2007morality}, we use the five-foundation theory of moral reasoning to understand our players' values and leanings. This theory does not give explicit definitions for the five foundations, but following recent work by \citet{ziems2022moral}, we can assume the following definition sketches:
\begin{enumerate}
    \item \textbf{Care:} wanting someone or something to be safe, healthy, and happy. \newline
    \textbf{Harm:} wanting someone or something to suffer physically, emotionally, socially, intellectually, or spiritually.
    \item \textbf{Fairness:} wanting to see individuals or groups treated equally or equitably \newline
    \textbf{Cheating:} wanting to see unfairness, injustice, bias, exclusion, or discrimination.
    \item \textbf{Loyalty:} wanting unity and seeing people keep promises or obligations to an in-group. \newline
    \textbf{Betrayal:} wanting to see people lie, abandon an in-group, or become isolated and divided.
    \item \textbf{Authority:} wanting to respect social roles, duties, privacy, peace, and order.  \newline
    \textbf{Subversion:} wanting to see people disrespect, disobey or cause disorder, challenge the status-quo, and do what they do not have permission to do.
    \item \textbf{Sanctity:} wanting people and things to be clean, pure, innocent, and holy. \newline
    \textbf{Degradation:} wanting people to follow selfish or crude desires and do things that make them or others dirty, corrupt, sick, repulsive, or perverted. 
\end{enumerate}

\paragraph{Moral Foundations Questionnaire} We use the associated \href{https://moralfoundations.org/questionnaires/}{Moral Foundations Questionnaire}, which we shortened to 12 questions as follows.

Please answer 12 questions about ``right'' and ``wrong.'' The prompts are the same in each case, but the considerations are different. 
\begin{enumerate}
    \item When you decide whether something is right or wrong, to what extent are the following considerations relevant to your thinking? Use the following scale: [0] Not at all relevant (It has nothing to do with my judgments of right and wrong); [1] Not very relevant; [2] Slightly relevant; [3] Somewhat relevant; [4] Very relevant; [5] Extremely relevant (It is one of the most important factors when I judge right and wrong)
    \begin{enumerate}
        \item Whether or not someone suffered emotionally.
        \item Whether or not some people were treated differently than others.
        \item Whether or not someone’s action showed love for his or her country.
        \item Whether or not someone showed a lack of respect for authority.
        \item Whether or not someone violated standards of purity and decency.
        \item Whether or not someone was good at math.
        \item Whether or not someone cared for someone weak or vulnerable.
        \item Whether or not someone acted unfairly.
        \item Whether or not someone did something to betray his or her group.
        \item Whether or not someone conformed to the traditions of society.
    \end{enumerate}
    \item Which of the following best describes your political views?
    \begin{enumerate}
        \item Liberal
        \item Moderate Liberal
        \item Moderate Conservative
        \item Conservative
        \item Libertarian
    \end{enumerate}
\end{enumerate}
\subsection{Instructions for Writing Rationales}
We explain that rationales should use at least 3 words to describe the connection between the clue and the target. Annotators were encouraged to be creative while trying to use one of the structures below. We imposed these structures for the sake of regularity.
\begin{enumerate}
    \item \texttt{MERONYM} x has y
    \begin{enumerate}
        \item a dog has a tail
        \item the pacific ocean has water
    \end{enumerate} 
    \item \texttt{HYPERNYM} x is a kind of y
    \begin{enumerate}
        \item bunkbed is a kind of bed
        \item whisper is a kind of communication
    \end{enumerate} 
    \item \texttt{SYNONYM} x means the same thing as y
    \begin{enumerate}
        \item car means the same thing as automobile
        \item sluggish means the same thing as slow
    \end{enumerate} 
    \item \texttt{ANTONYM} x means the opposite of y
    \begin{enumerate}
        \item civilian means the opposite of soldier
        \item fast means the opposite of slow
    \end{enumerate} 
    \item \texttt{ADJECTIVE} x describes y
    \begin{enumerate}
        \item brave describes a firefighter
        \item scary describes a clown
    \end{enumerate} 
    \item \texttt{AGENT} x does y
    \begin{enumerate}
        \item a star does twinkle
police do make an arrest
    \end{enumerate} 
    \item \texttt{CAUSE} x causes y
    \begin{enumerate}
        \item a bed causes people to sleep
        \item an oven causes food to bake
        \item a disease causes people to be sick
    \end{enumerate} 
    \item \texttt{PATIENT} x acts on y
    \begin{enumerate}
        \item a wrench acts on a bolt
        \item a doctor acts on a patient
    \end{enumerate} 
    \item \texttt{LOCATION} x has an environment y
    \begin{enumerate}
        \item a star has an environment firmament
    \end{enumerate} 
\end{enumerate}

\section{Training and Hyperparameters}

For our generation tasks, we perform use 5e-5 as our initial learning rate and perform a hyperparameter search over \{1...20\} epochs. For classification, we use the same splits and perform a hyperparameter sweep over learning rates (\{1e-4, 5e-4, 1e-5, 5e-5, 1e-6, 5e-6\}) and epochs (\{1...15\}). All models were trained on an NVIDIA A100 GPU. Across all experiments, GPU compute time was around 4-5 days.

\section{Artifact Details}
We use several models in our paper for their intended retrieval or generation task. Each model has its own license and number of parameters, listed below:
\begin{enumerate}
    \item T5~\citep{raffel2020exploring}, 220M parameters, is under the Apache 2.0 License.
    \item BART~\citep{lewis2019bart}, 140M, is under the Apache 2.0 License.
    \item fastText~\citep{bojanowski2016enriching} is under the MIT License.
    \item SentenceBERT~\citep{reimers-2020-multilingual-sentence-bert}, 33M variant, is under the Apache 2.0 License.
    \item BERT~\cite{devlin-etal-2019-bert} base, 110M, is under the Apache 2.0 License.
    \item XLNet~\cite{yang2019xlnet} base, 110M, is under the Apache 2.0 License.
    \item RoBERTAa~\cite{liu2019roberta} base, 123M, is under the Apache License 2.0.
\end{enumerate}
We plan on releasing \dataset{} and corresponding code under Creative Commons Attribution Share Alike 4.0 International. While our released dataset has extensive demographic information, we do not collect any identifiers that can uniquely isolate a person (e.g. name, MTurk ID, etc.)

\end{document}